\documentclass{article} 
\usepackage{nips13submit_e,times}
\usepackage{hyperref}
\usepackage{url}
\usepackage{microtype}
\usepackage{graphicx}
\usepackage{subfigure}
\usepackage{amsmath}
\usepackage{setspace}
\def\ie{{\em i.e.}}
\def\eg{{\em e.g.}}

\newcommand{\ignore}[1]{}

\nipsfinalcopy 

\begin{document}

\title{Robust and High Performance Face Detector}
\author{
  Yundong Zhang,\quad Xiang Xu,\quad Xiaotao Liu \\
  Vimicro AI Chip Technology Corporation, Beijing, China\\
  State Key Laboratory of Digital Multi-media Chip Technolgy, Beijing, China\\
  \texttt{\{raymond, xuxiang, liuxiaotao\}@vimicro.com}\\
}
\maketitle

\begin{abstract}
In recent years, face detection has experienced significant performance improvement with the boost of deep convolutional neural networks. In this report, we reimplement the state-of-the-art detector~\cite{DBLP:journals/corr/abs-1809-02693} and apply some tricks proposed in the recent literatures to obtain an extremely strong face detector, named VIM-FD. In specific, we exploit more powerful backbone network like DenseNet-121~\cite{DBLP:conf/cvpr/HuangLMW17}, revisit the data augmentation based on data-anchor-sampling proposed in~\cite{tang2018pyramidbox}, and use the max-in-out label and anchor matching strategy in~\cite{DBLP:conf/iccv/abs-1708-05237}. In addition, we also introduce the attention mechanism~\cite{wang2017fan,DBLP:conf/cvpr/ZhangQX0WY18} to provide additional supervision. Over the most popular and challenging face detection benchmark, \ie, WIDER FACE~\cite{DBLP:conf/cvpr/YangLLT16}, the proposed VIM-FD achieves state-of-the-art performance.
\end{abstract}

\section{Introduction}

Because face detection serves as a specific task for generic object detection, the development of generic object detection significantly promotes the development of face detection. Deep convolutional neural network (CNN) based object detectors have become more and more developed and achieved great success in recent years, owing to the significant progress of network architecture such as VGG~\cite{DBLP:journals/corr/SimonyanZ14a}, Inception~\cite{DBLP:conf/cvpr/SzegedyLJSRAEVR15,DBLP:conf/cvpr/SzegedyVISW16}, ResNet~\cite{DBLP:conf/cvpr/HeZRS16} and DenseNet~\cite{DBLP:conf/cvpr/HuangLMW17}. Advanced object detection frameworks can be divided into two categories: one-stage detector and two-stage detector. Most state-of-the-art methods use two-stage detectors, \eg, Faster R-CNN~\cite{DBLP:journals/pami/RenHG017}, R-FCN~\cite{DBLP:conf/nips/DaiLHS16}, FPN~\cite{DBLP:conf/cvpr/LinDGHHB17} and Cascade R-CNN~\cite{DBLP:journals/corr/abs-1712-00726}. These approaches first obtain a manageable number of region proposals called region of interest (RoI) and then pool out the corresponding features. In the second stage, R-CNN classifies and regresses each RoI again. By constract, one-stage detectors have the advantage of simple structures and high speed. SSD~\cite{DBLP:conf/eccv/LiuAESRFB16} and YOLO~\cite{DBLP:conf/cvpr/RedmonDGF16,DBLP:journals/corr/RedmonF16} have achieved good speed/accuracy trade-off. However, they can hardly surpass the accuracy of two-stage detectors. RetinaNet~\cite{DBLP:conf/iccv/LinPRK17} is the state-of-the-art one-stage detector that achieves comparable performance to two-stage detectors. It adopts an architecture modified from RPN~\cite{DBLP:conf/nips/RenHGS15} and focuses on addressing the class imbalance during training.

As a long-standing problem in computer vision, face detection has extensive applications including face alignment, face analysis, face recognition, etc. Starting from the pioneering work of Viola-Jones~\cite{DBLP:journals/ijcv/ViolaJ04}, face detection has also made great progress. The milestone work of Viola-Jones uses Haar feature and AdaBoost to train a cascade of face/non-face classifiers that achieves a good accuracy with real-time efficiency. After that, lots of works have focused on improving the performance with more sophisticated hand-crafted features~\cite{DBLP:journals/pami/LiaoJL16} and more powerful classifiers~\cite{DBLP:journals/ijcv/BrubakerWSMR08}. Besides the cascade structure,~\cite{DBLP:conf/eccv/MathiasBPG14} introduces deformable part models (DPM) into face detection tasks and achieves remarkable performance. However, these methods highly depend on the robustness of hand-crafted features and optimize each component separately, making face detection pipeline sub-optimal. Recent years have witnessed the advance of CNN-based face detectors. CascadeCNN~\cite{DBLP:conf/cvpr/LiLSBH15} develops a cascade architecture built on CNNs with powerful discriminative capability and high performance. Qin et al.~\cite{DBLP:conf/cvpr/QinYLH16} propose to jointly train CascadeCNN to realize end-to-end optimization. Faceness~\cite{DBLP:conf/iccv/YangLLT15} trains a series of CNNs for facial attribute recognition to detect partially occluded faces. MTCNN~\cite{DBLP:journals/spl/ZhangZLQ16} proposes to jointly solve face detection and alignment using several multi-task CNNs. UnitBox~\cite{DBLP:conf/mm/YuJWCH16} introduces a new intersection-over-union loss function.
The performances on several well-known datasets have been improved consistently, even tend to be saturated. 

In this paper, we reimplement the state-of-the-art detector~\cite{DBLP:journals/corr/abs-1809-02693} and utilize it as our baseline model. Then we revisit several proposed tricks from the following aspects: (1) Data augmentation method; (2) Matching and classification strategy; (3) Impact of the backbone network; (4) Attention mechanism in face detection. Through the organic integration of these tricks, we obtain a very powerful face detector, which achieves state-of-the-art result on WIDER FACE dataset.

\begin{figure*}[t!]
\centering
\includegraphics[width=0.9\linewidth]{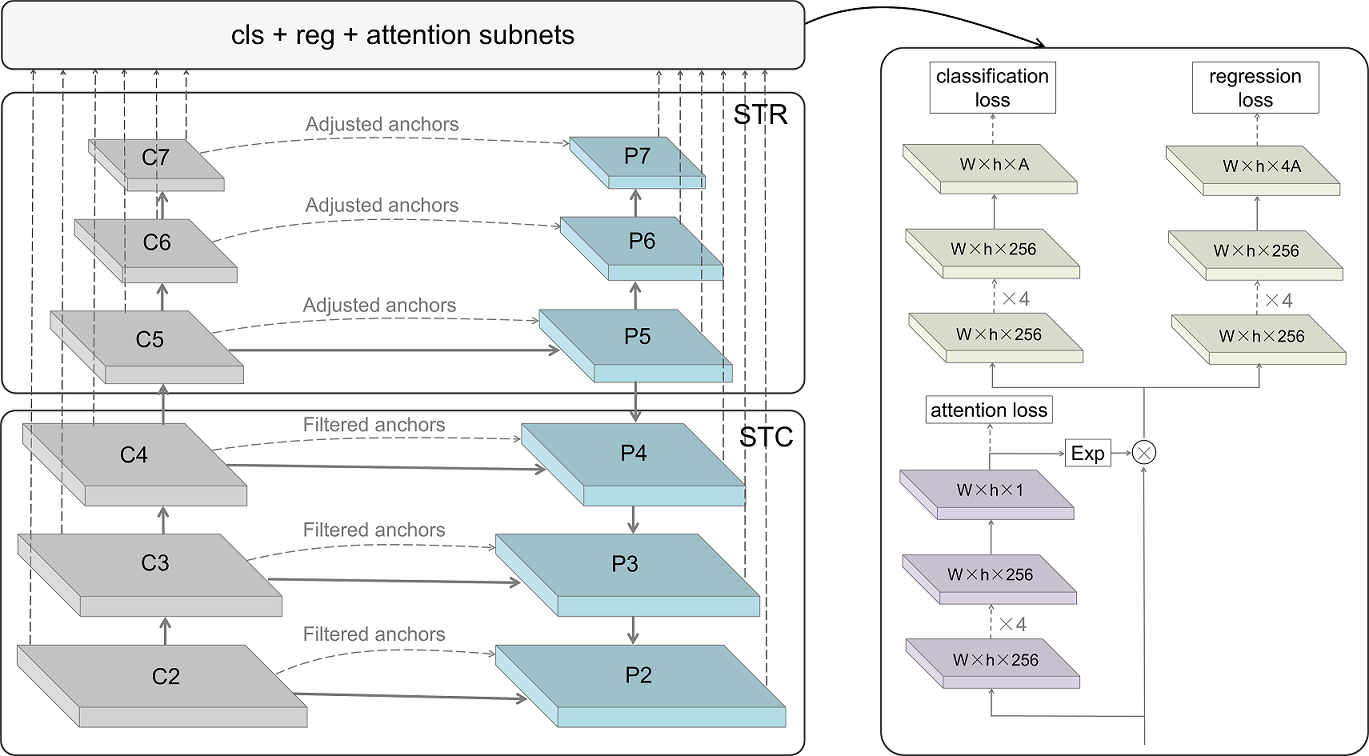}
\caption{Network structure of VIM-FD. It consists of STC, STR, and prediction subnets. STC uses the first-step classifier to filter out most simple negative anchors from low level detection layers to reduce the search space for the second-step classifier. STR applies the first-step regressor to coarsely adjust the locations and sizes of anchors from high level detection layers to provide better initialization for the second-step regressor. Prediction subnets perform classification, regression and attention map prediction jointly.}
\label{fig:framework}
\end{figure*}

\section{Related Work}
Face detection has been a challenging research field since its emergence in the 1990s. Viola and Jones pioneer to use Haar features and AdaBoost to train a face detector with promising accuracy and efficiency~\cite{DBLP:journals/ijcv/ViolaJ04}, which inspires several different approaches afterwards~\cite{DBLP:journals/pami/LiaoJL16,DBLP:journals/ijcv/BrubakerWSMR08}. Apart from those, another important work is the introduction of Deformable Part Model (DPM)~\cite{DBLP:conf/eccv/MathiasBPG14}. 

Recently, face detection has been dominated by the CNN-based methods. CascadeCNN~\cite{DBLP:conf/cvpr/LiLSBH15} improves detection accuracy by training a serious of interleaved CNN models and following work~\cite{DBLP:conf/cvpr/QinYLH16} proposes to jointly train the cascaded CNNs to realize end-to-end optimization. EMO~\cite{zhu2018seeing} proposes an Expected Max Overlapping score to evaluate the quality of anchor matching. SAFD~\cite{hao2017scale} develops a scale proposal stage which automatically normalizes face sizes prior to detection. S$^{2}$AP~\cite{song2018beyond} pays attention to specific scales in image pyramid and valid locations in each scales layer. PCN~\cite{shi2018real} proposes a cascade-style structure to rotate faces in a coarse-to-fine manner. Recent work~\cite{bai2018finding} designs a novel network to directly generate a clear super-resolution face from a blurry small one.

Additionally, face detection has inherited some achievements from generic object detectors, such as Faster R-CNN~\cite{DBLP:journals/pami/RenHG017}, SSD~\cite{DBLP:conf/eccv/LiuAESRFB16}, FPN~\cite{DBLP:conf/cvpr/LinDGHHB17}, RefineDet~\cite{DBLP:journals/corr/abs-1711-06897} and RetinaNet~\cite{DBLP:conf/iccv/LinPRK17}. Face R-CNN~\cite{wang2017face} combines Faster R-CNN with hard negative mining and achieves promising results. FaceBoxes~\cite{DBLP:conf/ijcb/abs-1708-05234} introduces a CPU real-time detecotor based on SSD. Face R-FCN~\cite{wang2017detecting} applies R-FCN in face detection and makes according improvements. The face detection model for finding tiny faces~\cite{DBLP:conf/cvpr/HuR17} trains separate detectors for different scales. S$^{3}$FD~\cite{DBLP:conf/iccv/abs-1708-05237} presents multiple strategies onto SSD to compensate for the matching problem of small faces. SSH~\cite{DBLP:conf/iccv/NajibiSCD17} models the context information by large filters on each prediction module. PyramidBox~\cite{tang2018pyramidbox} utilizes contextual information with improved SSD network structure. FAN~\cite{wang2017fan} proposes an anchor-level attention into RetinaNet to detect the occluded faces. 

\section{Proposed Approach}

The overall framework of VIM-FD is shown in Figure \ref{fig:framework}, we describe each component as follows.

\subsection{Backbone}
The original detector \cite{DBLP:journals/corr/abs-1809-02693} adopts ResNet-50~\cite{DBLP:conf/cvpr/HeZRS16} as the backbone network, which is a little bit obsolete currently. We explore several more powerful backbone network, such as ResNeXt~\cite{DBLP:journals/corr/XieGDTH16}, DenseNet~\cite{DBLP:conf/cvpr/HuangLMW17} and NASNet~\cite{DBLP:conf/cvpr/ZophVSL18}. Finally we adopt DenseNet-121 with 6-level feature pyramid structure as the backbone network for VIM-FD. The feature maps extracted from those four blocks are denoted as C2, C3, C4, and C5, respectively. C6 and C7 are just extracted by two simple down-sample $3\times3$ convolution layers after C5. The lateral structure between the bottom-up and the top-down pathways is the same as~\cite{DBLP:conf/cvpr/LinDGHHB17}. P2, P3, P4, and P5 are the feature maps extracted from lateral connections, corresponding to C2, C3, C4, and C5 that are respectively of the same spatial sizes, while P6 and P7 are just down-sampled by two $3\times3$ convolution layers after P5.

\subsection{STC and STR Module}
The STC module aims to filter out most simple negative anchors from low level detection layers to reduce the search space for the subsequent classifier, which selects C2, C3, C4, P2, P3, and P4 to perform two-step classification. While the STR module is designed to coarsely adjust the locations and sizes of anchors from high level detection layers to provide better initialization for the subsequent regressor, which selects C5, C6, C7, P5, P6, and P7 to conduct two-step regression. The loss function of these two modules is also same with~\cite{DBLP:journals/corr/abs-1809-02693}, which is described in detail as below:
\begin{equation}
\begin{aligned}
{\cal L}_\text{STC} (\{p_i\},\{q_i\})=\frac{1}{N_{\text{s}_1}}  \sum_{i\in \Omega}{\cal L}_{\text{FL}}(p_i,l_i^\ast) \\
+ \frac{1}{N_{\text{s}_2}}  \sum_{i\in \Phi}{\cal L}_{\text{FL}}(q_i, l_i^\ast),
\end{aligned}
\end{equation}
where $i$ is the index of anchor in a mini-batch, $p_i$ and $q_i$ are the predicted confidence of the anchor $i$ 
being a face in the first and second steps, $l_i^\ast$ is the ground truth class label of anchor $i$, $N_{\text{s1}}$ and $N_{\text{s2}}$ are the numbers of positive anchors in the first and second steps, $\Omega$ represents a collection of samples selected for two-step classification, and $\Phi$ represents a sample set that remains after the first step filtering. The binary classification loss ${\cal L}_{\text{FL}}$ is the sigmoid focal loss over two classes (face {\em vs.} background).
\begin{equation}
\begin{aligned}
{\cal L}_\text{STR}(\{x_i\},\{t_i\})=\sum_{i\in \Psi}[l_i^\ast=1]{\cal L}_{\text{r}}(x_i, g_i^\ast) \\
+ \sum_{i\in \Phi}[l_i^\ast=1]{\cal L}_{\text{r}}(t_i, g_i^\ast),
\end{aligned}
\end{equation}
where $g_i^\ast$ is the ground truth location and size of anchor $i$, $x_i$ is the refined coordinates of the anchor $i$ in the first step, $t_i$ is the coordinates of the bounding box in the second step, $\Psi$ represents a collection of samples selected for two-step regression, $l_i^\ast$ and $\Phi$ are the same as defined in STC. Similar to Faster R-CNN \cite{DBLP:journals/pami/RenHG017}, we use the smooth L$_1$ loss as the regression loss $L_{\text{r}}$. The Iverson bracket indicator function $[l_i^\ast=1]$ outputs $1$ when the condition is true, \ie, $l_i^\ast=1$ (the anchor is not the negative), and $0$ otherwise. Hence $[l_i^\ast=1]{\cal L}_{\text{r}}$ indicates that the regression loss is ignored for negative anchors.

\subsection{Attention module}
Recently, attention mechanism is applied continually in object detection and face detection. DES~\cite{DBLP:conf/cvpr/ZhangQX0WY18} utilizes the idea of weakly supervised semantic segmentation, to provide high semantic meaningful and class-aware features to activate and calibrate feature map used in the object detection. FAN~\cite{wang2017fan} introduces anchor-level attention to highlight the features from the facial regions and successfully relieving the risk from the false positives.

We apply the attention subnet in FAN to the P2, P3, P4, P5, P6 and P7 layers, the specific structure of which is shown in Figure \ref{fig:framework}. Specifically, the attention supervision information is obtained by filling the ground-truth box. And the supervised masks are associated to the ground-truth faces assigned to the anchors in the current detection layer. Because the first step and the second step share the same detection subnet, we also apply attention subnet on bottom-up levels, but we do not calculate attention loss on these layers. We define the attention loss function as:
\begin{equation}
\begin{aligned}
{\cal L}_\text{ATT}(\{m_i\})=\sum_{i\in \mathrm{X}}{\cal L}_{\text{sig}}(m_i, m_i^\ast),
\end{aligned}
\end{equation}
where $m_i$ is the predicted attention map generated per level in the second step, $m_i^\ast$ is the ground truth attention mask of the $i$ th detection layer, $\mathrm{X}$ represents the set of detection layers applied to attention mechanism (\ie, P2, P3, P4, P5, P6 and P7), and $L_{\text{sig}}$ is pixel-wise sigmoid cross entropy loss. 

\subsection{Max-in-out Label}
S$^3$FD~\cite{DBLP:conf/iccv/abs-1708-05237} introduces max-out background label to reduce the false positives of small negatives. PyramidBox~\cite{tang2018pyramidbox} uses this strategy on both positive and negative samples. In specific, this strategy first predicts $c_p + c_n$ scores for each prediction module, and then selects $\max{c_p}$ as the positive score. Similarly, it chooses the max score of $c_n$ to be the negative score. In our VIM-FD, we employ the max-in-out label in the classification subnet, and set $c_p = 3$ and $c_n = 3$ to recall more faces and reduce false positives simultaneously.

\subsection{Anchor Design and Matching}
The design of anchor scale and ratio keeps same with \cite{DBLP:journals/corr/abs-1809-02693}. At each pyramid level, we use two specific scales of anchors as same as \cite{DBLP:journals/corr/abs-1809-02693} (\ie, $2S$ and $2\sqrt{2}S$, where $S$ represents the total stride size of each pyramid level) and one aspect ratios (\ie, $1.25$). In total, there are $A=2$ anchors per level and they cover the scale range $8-362$ pixels across levels with respect to the network's input image.

During the training phase, anchors need to be divided into positive and negative samples. Specifically, anchors are assigned to ground-truth face boxes using an intersection-over-union (IoU) threshold of $\theta_{p}$; and to background if their IoU is in $[0, \theta_{n})$. If an anchor is unassigned, which may happen with overlap in $[\theta_{n}, \theta_{p})$, it is ignored during training. Empirically, we set $\theta_{n}=0.3$ and $\theta_{p}=0.7$ for the first step, and $\theta_{n}=\theta_{p}=0.35$ for the second step. This setting draws on the scale compensation anchor matching strategy in S$^3$FD~\cite{DBLP:conf/iccv/abs-1708-05237}, aiming to improve the recall rate of small faces. The setting is based on the observation that faces whose scale distribute away from anchor scales can not match enough anchors. To solve this issue, we decrease the IoU threshold to increase the average number of matched anchors. The scale compensation anchor matching strategy greatly increases the matched anchors of tiny and outer faces, which notably improves the recall rate of these faces.

\subsection{Data Augmentation}
We employ the data-anchor-sampling method in PyramidBox~\cite{tang2018pyramidbox} to diversify the scale distribution of training samples and construct a robust model. Specifically, we first randomly select a face of size $S_{face}$ in a batch. Let
\begin{equation}
\begin{aligned}
i_{anchor} = \underset{i}{\arg\min} \ \mbox{abs}(S_{anchor_i} - S_{face})
\end{aligned}
\end{equation}
be the index of the nearest anchor scale from the selected face, then we choose a random index $i_{target}$ in the set $\{ 0, 1, \ldots, \mbox{min}(5, i_{anchor} + 1)\}$, thus we get the image resize scale
\begin{equation}
\begin{aligned}
S^* = random(S_{i_{target}}/2, S_{i_{target}}*2) / S_{face}.
\end{aligned}
\end{equation}
By resizing the original image with the scale $S^*$ and cropping a standard size of $640\times640$ containing the selected face randomly, we get the anchor-sampled training data.

\subsection{Loss Function}
We append a hybrid loss at the end of the deep architecture to jointly optimize model parameters, which is just the sum of the STC loss, the STR loss and the ATT loss:

\begin{equation}
\begin{aligned}
{\cal L}={\cal L}_\text{STC} + {\cal L}_\text{STR} + {\cal L}_\text{ATT},
\end{aligned}
\end{equation}

\begin{figure*}[t]
\centering
\subfigure[Val: Easy]{
\label{fig:ve}
\includegraphics[width=0.49\linewidth]{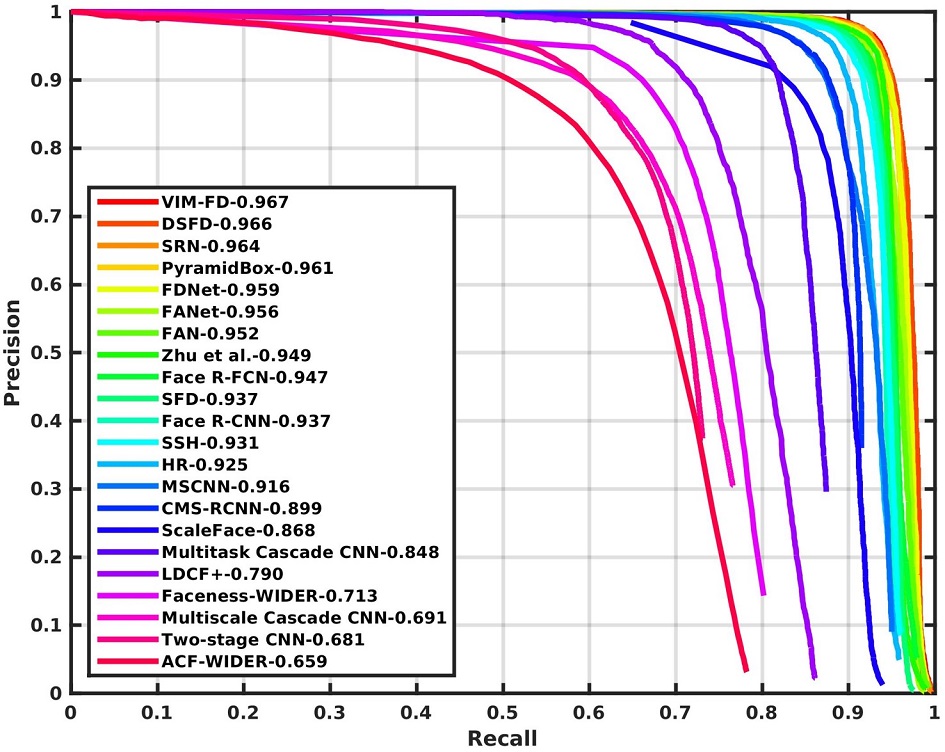}}
\subfigure[Test: Easy]{
\label{fig:te}
\includegraphics[width=0.49\linewidth]{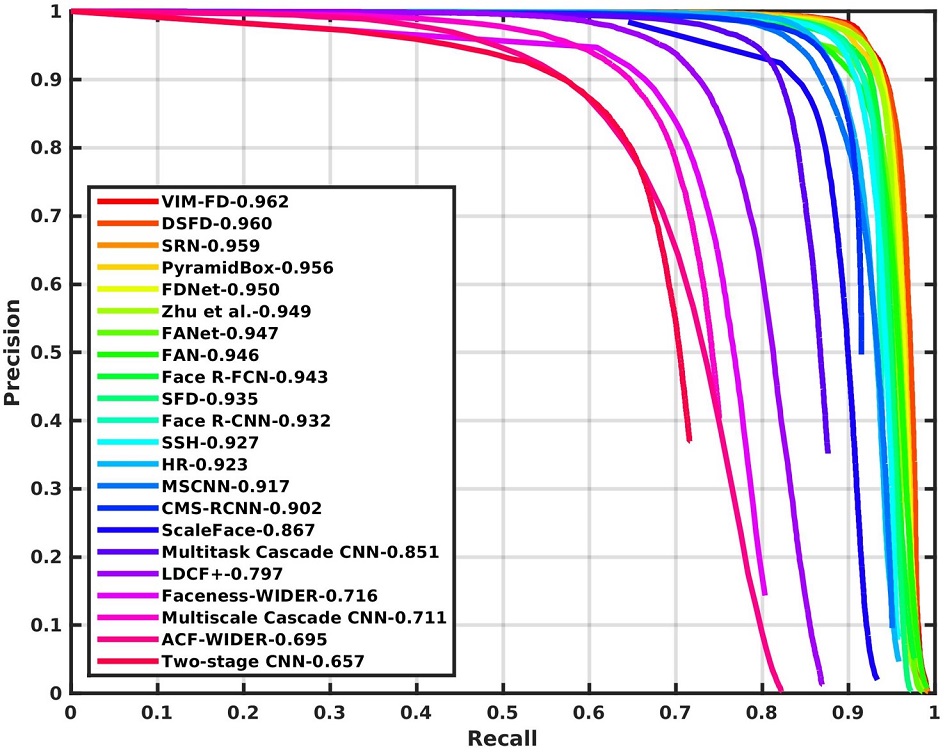}}
\subfigure[Val: Medium]{
\label{fig:vm}
\includegraphics[width=0.49\linewidth]{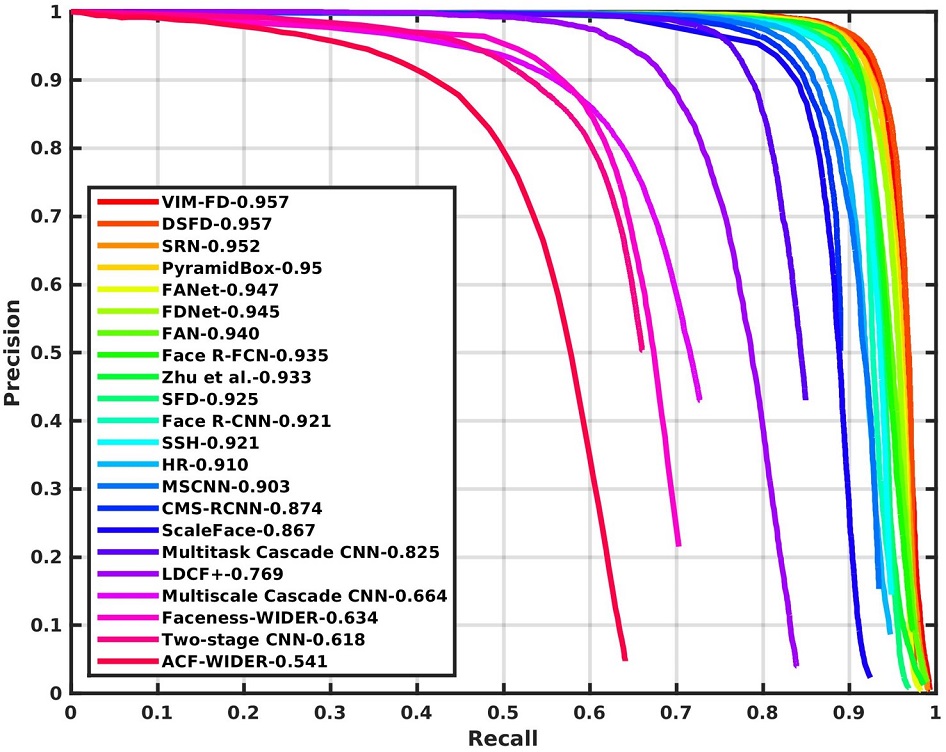}}
\subfigure[Test: Medium]{
\label{fig:tm}
\includegraphics[width=0.49\linewidth]{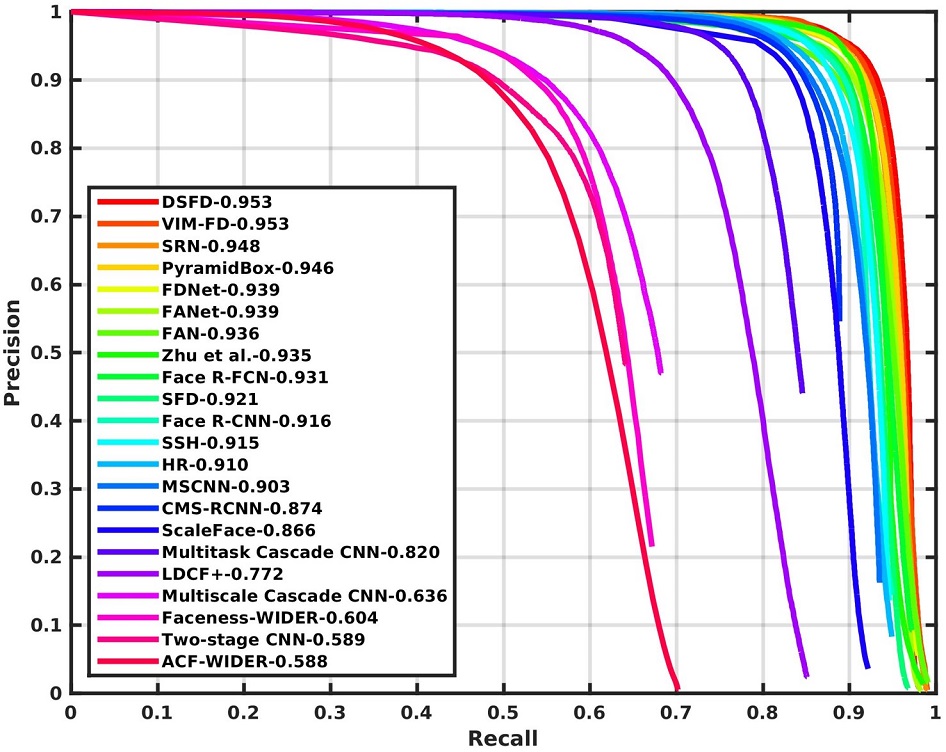}}
\subfigure[Val: Hard]{
\label{fig:vh}
\includegraphics[width=0.49\linewidth]{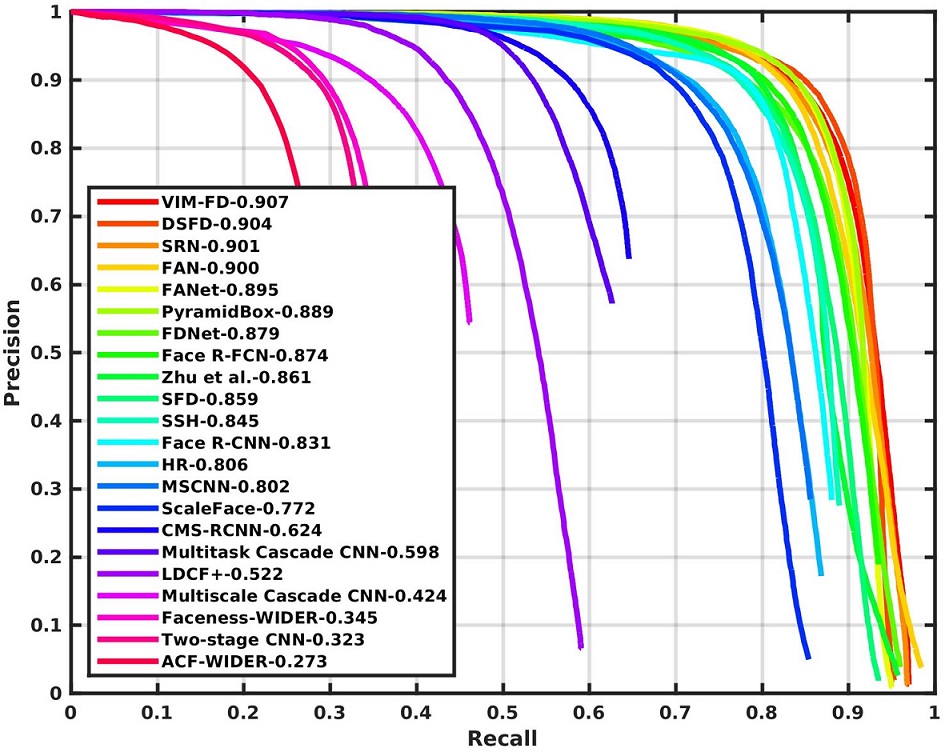}}
\subfigure[Test: Hard]{
\label{fig:th}
\includegraphics[width=0.49\linewidth]{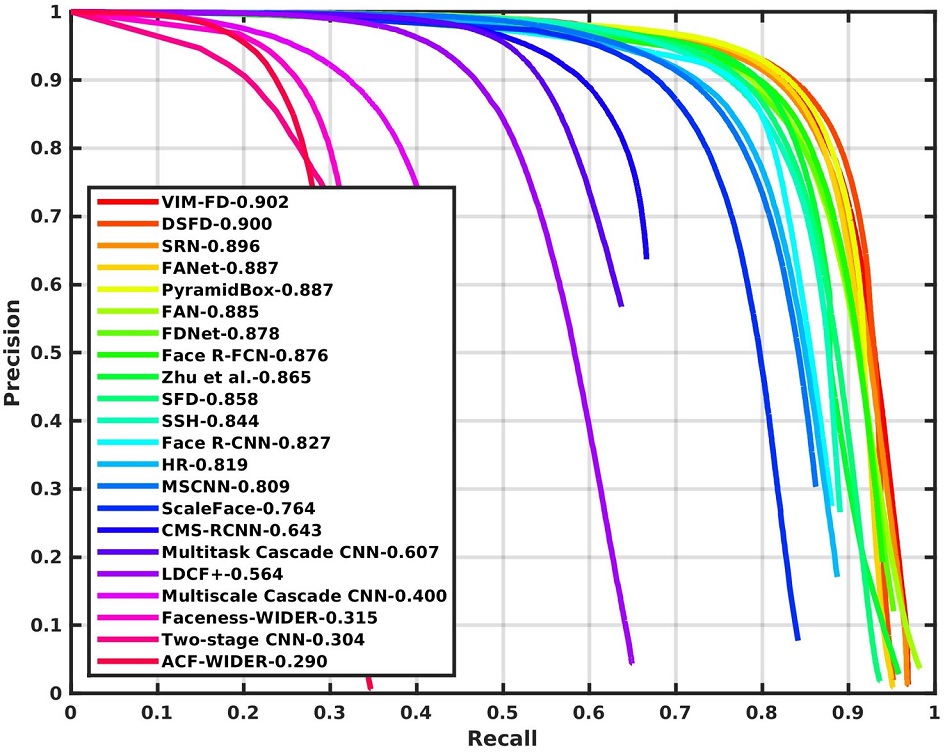}}
\vspace{-5mm}
\caption{Precision-recall curves on WIDER FACE validation and testing subsets.}
\vspace{-5mm}
\label{fig:wider-face-ap}
\end{figure*}

\section{Experiments}
The backbone network is initialized by the pretrained DenseNet-121 model~\cite{DBLP:journals/ijcv/RussakovskyDSKS15} and all the parameters in the newly added convolution layers are initialized by the ``xavier'' method. We fine-tune the model using SGD with $0.9$ momentum, $0.0001$ weight decay, and batch size $32$. We set the learning rate to $10^{-2}$ for the first $100$ epochs, and decay it to $10^{-3}$ and $10^{-4}$ for another $20$ and $10$ epochs, respectively. We implement VIM-FD using the PyTorch library~\cite{paszke2017pytorch}.

\subsection{Dataset}
The WIDER FACE dataset~\cite{DBLP:conf/cvpr/YangLLT16} consists of $393,703$ annotated face bounding boxes in $32,203$ images with variations in pose, scale, facial expression, occlusion, and lighting condition. The dataset is split into the training ($40\%$), validation ($10\%$) and testing ($50\%$) sets, and defines three levels of difficulty: Easy, Medium, Hard, based on the detection rate of EdgeBox~\cite{DBLP:conf/eccv/ZitnickD14}. Due to the variability of scale, pose and occlusion, WIDER FACE dataset is one of the most challenge face datasets. Our VIM-FD is trained only on the training set and evaluated on both validation set and testing set.

\subsection{Experimental Results}
\begin{figure}[h]
\centering
\includegraphics[width=1.0\textwidth]{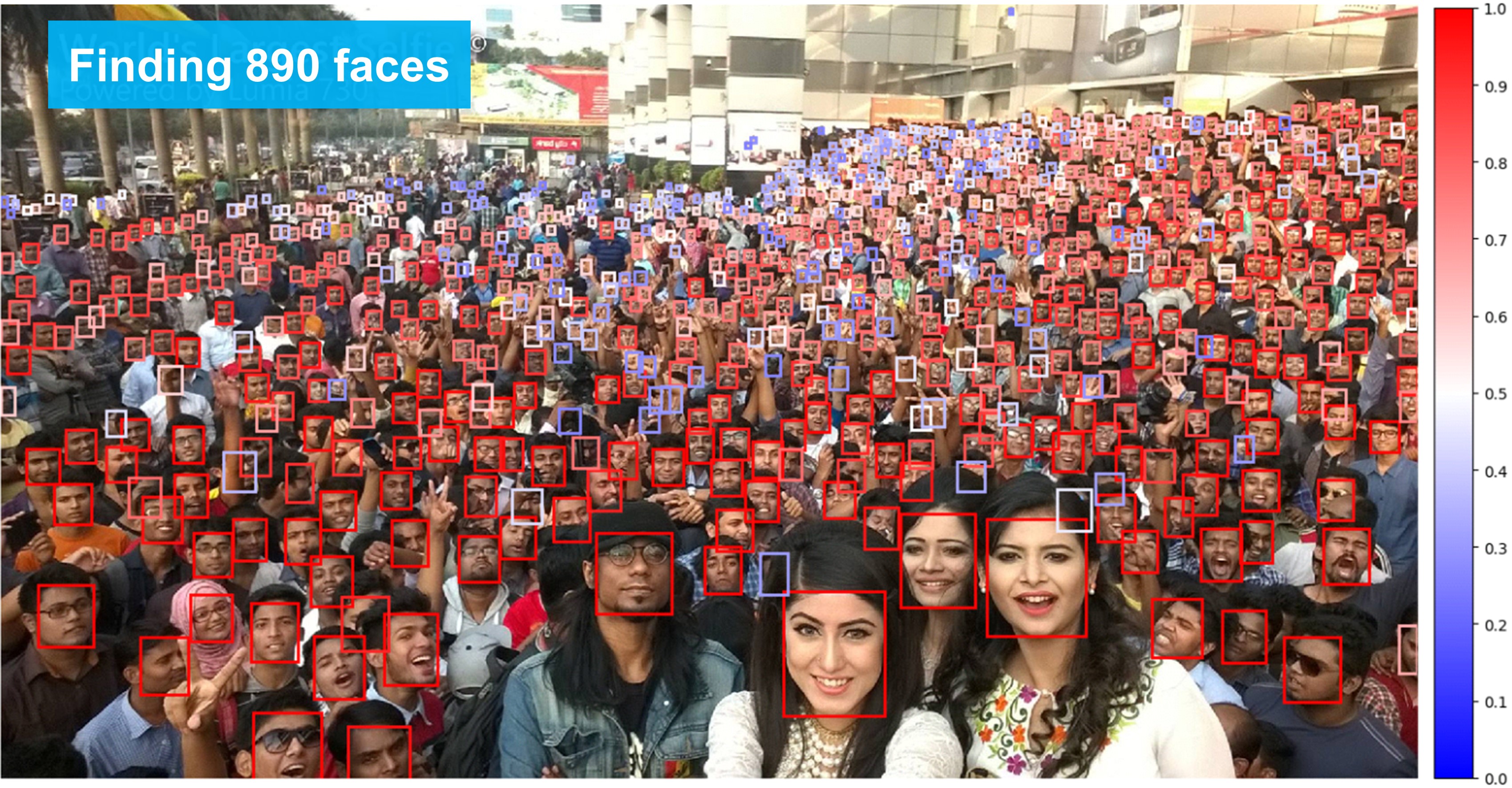}
\caption{Impressive qualitative result. VIM-FD finds $890$ faces out of the reported $1000$ faces. The confidences of the detections are presented in the color bar on the right hand. Best viewed in color.}
\label{fig:slumia}
\end{figure}

\begin{figure}[!h]
\centering
\subfigure[Scale attribute. Our VIM-FD is able to detect faces at a continuous range of scales.]{
\label{fig:scale}
\includegraphics[width=1.0\linewidth]{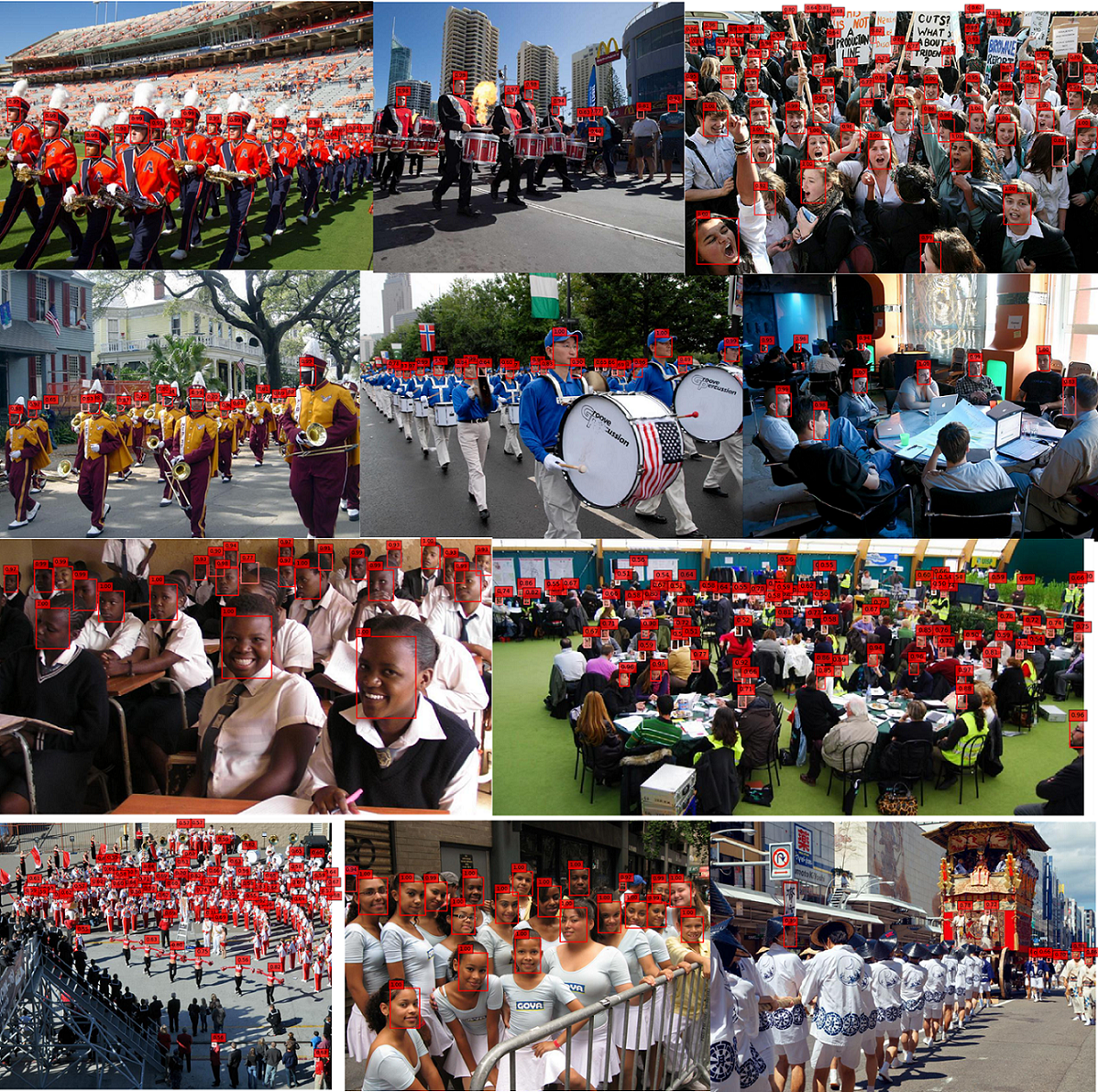}}
\subfigure[Our VIM-FD is robust to blur, expression, illumination, makeup, occlusion and pose.]{
\label{fig:widerface}
\includegraphics[width=1.0\linewidth]{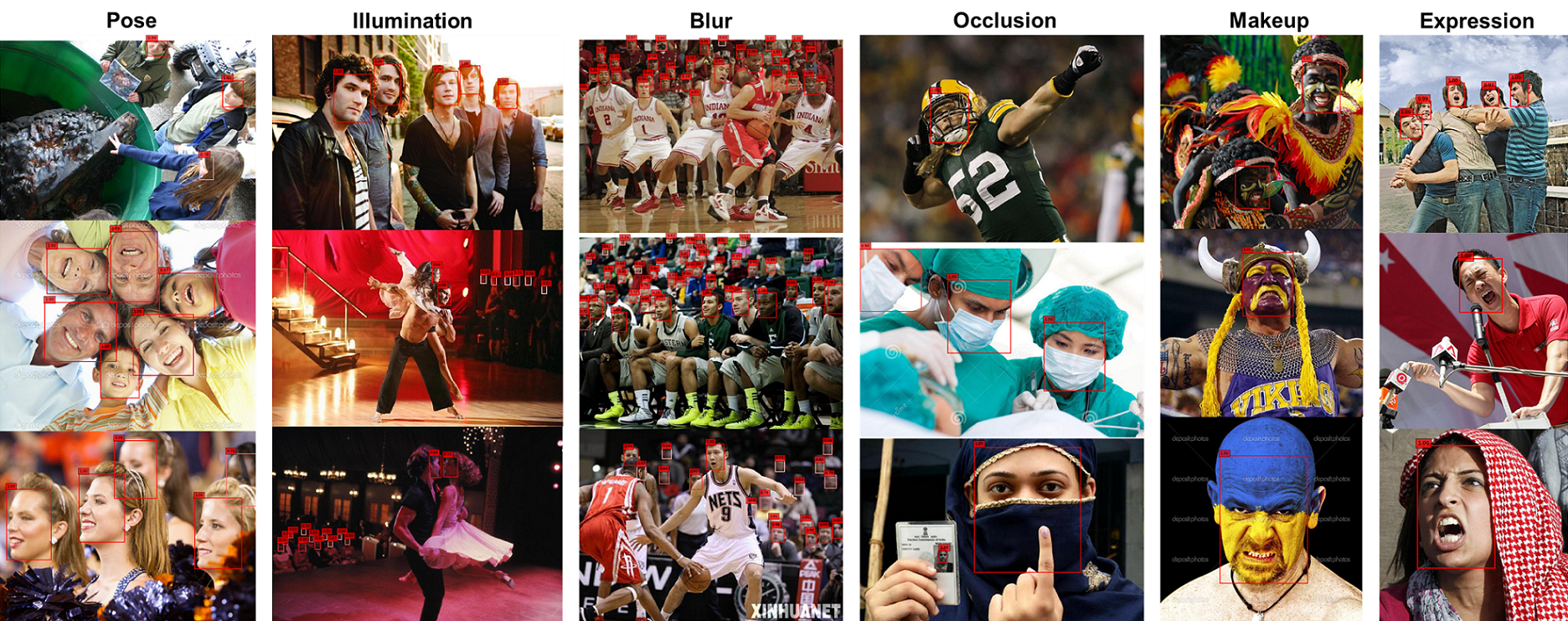}}
\caption{Qualitative results on WIDER FACE. We visualize some examples for each attribute. Please zoom in to see small detections.}
\label{fig:widerface_xg}
\end{figure}

We compare VIM-FD with twenty-one state-of-the-art face detection methods on both the validation and testing sets. To obtain the evaluation results on the testing set, we submit the detection results of VIM-FD to the authors for evaluation. As shown in Figure \ref{fig:wider-face-ap}, we find that VIM-FD performs favourably against the state-of-the-art based on the average precision (AP) across the three subsets, especially on the Hard subset which contains a large amount of small faces. Specifically, it produces the best AP scores in all subsets of both validation and testing sets, \ie, $96.7\%$ (Easy), $95.7\%$ (Medium) and $90.7\%$ (Hard) for validation set, and $96.2\%$ (Easy), $95.3\%$ (Medium) and $90.2\%$ (Hard) for testing set. Except that the AP of Medium subset is equal to DSFD~\cite{DBLP:journals/corr/abs-1810-10220}, our results surpass all approaches, which demonstrates the superiority of the proposed detector. We first show an impressive qualitative result of the World Largest Selfie\footnote{\url{https://www.cs.cmu.edu/~peiyunh/tiny/}} in Figure \ref{fig:slumia} and VIM-FD successfully finds $890$ faces out of the reported $1000$ faces. We also demonstrate some qualitative results on WIDER FACE in Figure \ref{fig:widerface_xg}, indicating the proposed VIM-FD is robust to scale, blur, expression, illumination, makeup, occlusion and pose.

\section{Conclusion}
In this report, we reimplement the state-of-the-art detector~\cite{DBLP:journals/corr/abs-1809-02693} and revisit several tricks proposed in the recent literatures to obtain an extremely strong face detector, named VIM-FD. In specific, we make some explorations in the following aspects: (1) Data augmentation method; (2) Matching and classification strategy; (3) Impact of the backbone network; (4) Attention mechanism in face detection. Extensive experiments on the WIDER FACE dataset demonstrate that VIM-FD achieves the state-of-the-art detection performance.

\vspace{-5mm}

\bibliographystyle{unsrt}
\bibliography{reference}
\end{document}